\documentclass[manuscript,screen,nonacm]{acmart}%

\AtBeginDocument{%
  \providecommand\BibTeX{{%
    \normalfont B\kern-0.5em{\scshape i\kern-0.25em b}\kern-0.8em\TeX}}}

\setcopyright{none}%

\usepackage{tasks}
\usepackage{todonotes}

\usepackage{graphicx}
\usepackage{caption}
\usepackage{subcaption}
\usepackage{wrapfig}

\usepackage{booktabs}

\usepackage{enumitem}

\usepackage{fontawesome}

\usepackage[framemethod=tikz]{mdframed}
\definecolor{myblue}{rgb}{0.122, 0.435, 0.698}
\definecolor{mygreen}{rgb}{0.125, 0.525, 0.220}
\definecolor{myyellow}{rgb}{0.588, 0.439, 0.000}
\newmdenv[innerlinewidth=0.5pt,roundcorner=4pt,innerleftmargin=6pt,
          innerrightmargin=6pt,innertopmargin=6pt,innerbottommargin=6pt,
          linecolor=myblue,backgroundcolor=myblue!25!white]{mybluebox}
\newmdenv[innerlinewidth=0.5pt,roundcorner=4pt,innerleftmargin=6pt,
          innerrightmargin=6pt,innertopmargin=6pt,innerbottommargin=6pt,
          linecolor=mygreen,backgroundcolor=mygreen!25!white]{mygreenbox}
\newmdenv[innerlinewidth=0.5pt,roundcorner=4pt,innerleftmargin=6pt,
          innerrightmargin=6pt,innertopmargin=6pt,innerbottommargin=6pt,
          linecolor=myyellow,backgroundcolor=myyellow!25!white]{myyellowbox}

\begin{document}

\title[%
Simply Logical (Fully Interactive Online Edition)%
]{%
Simply Logical -- %
Intelligent Reasoning by Example\\ %
(Fully Interactive Online Edition)
}

\author{Peter Flach}
\email{Peter.Flach@bristol.ac.uk}
\orcid{0000-0001-6857-5810}
\affiliation{%
  \institution{Intelligent Systems Laboratory, University of Bristol}
  \country{United Kingdom}
}

\author{Kacper Sokol}
\email{K.Sokol@bristol.ac.uk}
\orcid{0000-0002-9869-5896}
\affiliation{%
  \institution{Intelligent Systems Laboratory, University of Bristol}
  \country{United Kingdom}
}

\begin{abstract}
\begin{mygreenbox}
\begin{description}[labelindent=0em,%
                    labelwidth=6em,
                    labelsep=1em,
                    itemindent=0em,
                    leftmargin=7em]
    \item [Online Book] \url{https://book.simply-logical.space/}%
    \item [Book Source] \url{https://github.com/simply-logical/simply-logical/}%
\end{description}
\end{mygreenbox}
\end{abstract}


\maketitle

\setcounter{secnumdepth}{0}

\section{Evolution of The Book}%

``Simply Logical -- Intelligent Reasoning by Example'' by Peter Flach was first published by John Wiley in 1994~\cite{flach1994simply}. %
It could be purchased as book-only or with a 3.5 inch diskette containing the SWI-Prolog~\cite{wielemaker2012swiprolog} programmes printed in the book
(for various operating systems). %
In 2007 the copyright reverted back to the author at which point the book and programmes were made freely available online; %
the print version is no longer distributed through John Wiley publishers. %
In 2015, as a pilot, we ported most of the original book into an online, interactive website using SWI-Prolog's SWISH platform~\cite{wielemaker2015swish}. %
Since then, we launched the \emph{Simply Logical} open source organisation\footnote{\url{https://simply-logical.space/}} committed to maintaining a suite of freely available interactive online educational resources\footnote{\url{https://github.com/simply-logical/}} about Artificial Intelligence and Logic Programming with Prolog. %
With the advent of new educational technologies we were inspired to rebuild the book from the ground up using the Jupyter Book platform~\cite{jupyter2020book} enhanced with a collection of bespoke plugins that implement, among other things, interactive SWI-Prolog code blocks that can be executed directly in a web browser. %
This new version is more modular, easier to maintain, and can be split into custom teaching modules, in addition to being modern-looking, visually appealing, and compatible with a range of (mobile) devices of varying screen sizes.%

\section{Artificial Intelligence and Logic Programming with Prolog}

The book discusses methods to implement intelligent reasoning by means of Prolog programmes. %
It is written from the shared viewpoints of \emph{Computational Logic} -- which strives to automate various kinds of reasoning -- and \emph{Artificial Intelligence} -- which seeks to implement aspects of intelligent behaviour as computation. %
The combination of these two perspectives distinguishes this book from among its peers and offers a unique learning experience. %
The readers working in Artificial Intelligence will find a detailed treatment of how the power of logic can be harnessed to solve some of the (practical) problems they may be facing. %
To support this learning objective the book provides a variety of interactive code examples in the domains of natural language interpretation, abductive and inductive reasoning, and reasoning by default. %
Those acquainted with Logic Programming, on the other hand, will be interested in the practical side of many topics that in other educational materials mostly receive theoretical treatment -- many advanced programmes presented and explained in this book are not, in a didactic form, available elsewhere. %
The readers unfamiliar with either field will benefit from a comprehensive learning resource that collects diverse subjects and presents them consistently across accessible and engaging modules. %
This approach offers the best of both worlds: it introduces the theory of Logic Programming that does not intimidate novices to the field with superfluous mathematical machinery, nonetheless it offers strong theoretical foundations for the practical aspects of programming for Artificial Intelligence.%

The book consists of three parts. %
Part I deals with Logic and Logic Programming and covers:%
\begin{itemize}
    \item key concepts in Logic Programming, such as programme clauses, query answering, proof trees, and recursive data structures (Chapter 1);%
    \item a rigorous discussion of resolution theorem proving in clausal logic, stepping through propositional clausal logic, relational clausal logic (without functors), full clausal logic, and definite clause logic, on the way dealing with concepts such as Herbrand models and resolution refutations, as well as meta-theoretical notions like soundness and completeness (Chapter 2); and%
    \item practical aspects of Prolog programming, including SLD-trees, cuts, arithmetic expressions, second-order predicates (e.g., \texttt{setof}), various programming techniques (e.g., accumulators and difference lists) as well as a general programming methodology and meta-interpreters (Chapter 3).%
\end{itemize}

Part II shifts the focus from the Logic Programming perspective to the Artificial Intelligence viewpoint, predominantly dealing with graphs and search. %
Chapter 4 discusses graphs found naturally in Prolog: in the form of trees represented by terms (e.g., parse trees) and as search spaces spanned by predicates (e.g., SLD-trees). %
Next, Chapter 5 overviews depth-first search, iterative deepening, and breadth-first search in the context of Logic Programming; %
moreover, it develops a breadth-first Prolog meta-interpreter and an (inefficient) interpreter for full clausal logic, in addition to illustrating forward chaining with a programme that generates Herbrand models of a set of clauses. %
Chapter 6 discusses best-first search and its optimality, leading to the A\(^\star\) algorithm and a brief discussion of non-exhaustive heuristic search strategies (beam search and hill-climbing).%

Finally, Part III encompasses advanced topics. %
Chapter 7 covers natural language parsing and interpretation, in particular definite clause grammars. %
It also looks at semantics and natural language generation, culminating in a simple question-answering agent. %
Chapter 8 discusses reasoning with incomplete information in view of the Closed World Assumption and Predicate Completion. %
Specifically, it introduces abductive reasoning as well as default reasoning by means of negation as failure and defeasible default rules. %
Then, Chapter 9 deals with inductively inferring a logic programme from examples, introducing and implementing concepts such as generality between clauses and anti-unification, which are fundamental to Inductive Logic Programming.%

Appendices offer answers to selected exercises, an overview of built-in predicates in Prolog, a library of utility predicates underpinning many programmes introduced throughout the book, and listings of two larger programmes for transforming a Predicate Logic formula to clausal logic and performing Predicate Completion.%

As suggested by the title, this book presents intelligent reasoning techniques by example, therefore every method is accompanied by a Prolog implementation that, in the online edition, can be executed directly in a web browser. %
These code listings serve two didactic purposes: they engender understanding through hands-on experience, but more importantly the declarative reading of each implementation constitutes an integral part of the explanation of the underlying technique. %
The more elaborate programmes are carefully designed to explicitly communicate the steps taken to achieve the goal, explaining key issues along the way. %
The book therefore does not just focus on ``How is this done in Prolog?'', but rather on ``How should I solve this problem, were I to start from scratch?'' In other words, it embodies the ``teaching by showing, learning by doing'' philosophy. %
Despite the strong focus on practical examples, a substantial portion of the book is devoted to the theoretical underpinnings of clausal logic and Logic Programming, however their presentation is primarily motivated by their instrumental role in understanding and implementing each particular technique.%
\footnote{The book's preface offers a more comprehensive explanation of its instructional design and content: \url{https://book.simply-logical.space/src/simply-logical.html\#author-s-preface}.}%

The book has been the foundation of multiple courses taught to undergraduate and master's students at the University of Bristol and elsewhere, e.g., King's College London, Queen Mary University of London, University of Cardiff, University College Cork, Tufts University, St. Joseph's University, University of Freiburg, Uppsala University, Free University Brussels, and University of Sevilla\footnote{\url{http://people.cs.bris.ac.uk/~flach/SL/Courses.html}}. %
Specifically, the \emph{Artificial Intelligence with Logic Programming} undergraduate course\footnote{\url{https://coms30106.github.io/}} offered at the University of Bristol follows the book quite closely; %
a collection of slides whose structure mirrors the presentation of material in the book is freely available as a supplementary educational resource\footnote{\url{https://github.com/simply-logical/slides/}}. %
A spiritual successor of this unit is \emph{Computational Logic for Artificial Intelligence} now taught as part of the curriculum delivered by the Interactive Artificial Intelligence Centre for Doctoral Training\footnote{\url{https://www.bristol.ac.uk/cdt/interactive-ai/}} (IAI CDT). %
This unit takes a more modern approach by discussing the interconnections between logic programming (with SWI-Prolog) and recent developments in data-driven machine learning\footnote{\url{https://github.com/simply-logical/ComputationalLogic/}}.%

\section{The Online Edition}%

This online book offers an intuitive and comprehensive introduction to \emph{Artificial Intelligence and Logic Programming} in the form of an accessible educational resource with a modern look and feel. %
The learning experience is enhanced with \emph{exercises} and \emph{practical coding examples}; %
the latter are delivered through interactive code boxes that can be executed directly in a web browser, thus fulfilling the original vision for the paperback edition that was curtailed by the lack of appropriate technologies at the time of its publication. %
It is written with three kinds of readers in mind: %
\emph{Artificial Intelligence} researchers or practitioners; %
\emph{Logic Programming} researchers or practitioners; and %
students (advanced undergraduate or graduate level) in both fields. %
The book can be used as a teaching aid, but it is also suitable for self-study. %
It can be adopted as is or the content can be rearranged into a bespoke learning resource, which is made possible by the modular design of the book source that is split into a collection of files based on their type:%
\begin{itemize}
    \item text is written in Markdown -- each sub-section is placed in a separate file;%
    \item code is released as SWI-Prolog scripts; and%
    \item figures are available as Scalable Vector Graphics (SVG).%
\end{itemize}
All of these are published on GitHub\footnote{\url{https://github.com/simply-logical/simply-logical/}}, which allows to easily reuse these materials, incorporate them into bespoke courses, or adapt them into alternative educational resources such as practical training sessions.%

Having used the book for undergraduate and postgraduate teaching for many years, we decided against making major structural changes or adding substantial new material. %
We have, of course, taken the opportunity to include small improvements and corrections, such as using anonymous variables wherever appropriate, and slight punctuation improvements in clauses and queries. %
Other enhancements include additional examples collected during years of teaching, formatted in clearly identifiable boxes.%

\begin{figure}[t]
    \includegraphics[width=\textwidth]{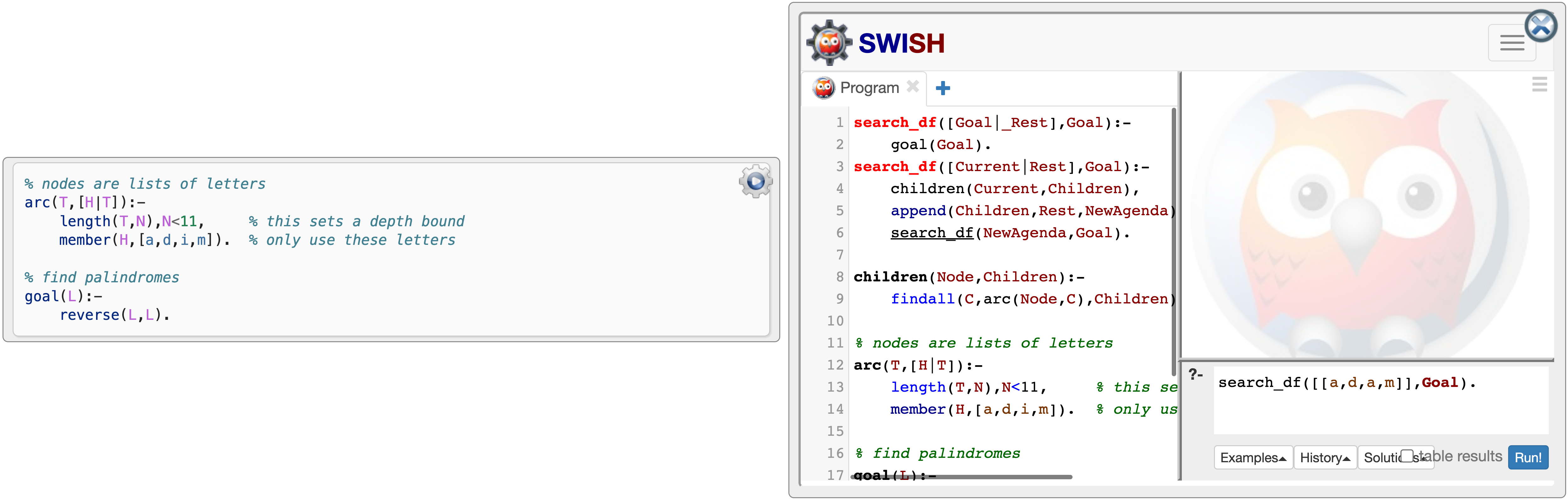}%
    \caption{Interactive SWI-Prolog code box based on SWISH. A code listing (left) can be turned into an interactive code box (right) by pressing the ``play'' button placed in the top-right corner.\label{fig:codebox}}
\end{figure}

From a technical perspective, the development of the online edition required us to implement a collection of Jupyter Book plugins (that also work with Sphinx) spanning functionality specific to SWI-Prolog\footnote{\url{https://www.swi-prolog.org/}}~\cite{wielemaker2012swiprolog}, later extended to be compatible with the cplint\footnote{\url{https://cplint.ml.unife.it/}}~\cite{riguzzi2018foundations} and ProbLog\footnote{\url{https://dtai.cs.kuleuven.be/problog/}}~\cite{deraedt2007problog} programming languages. %
Specifically, we released \texttt{sphinx-prolog}\footnote{\url{https://github.com/simply-logical/sphinx-prolog/}}, which allows to embed interactive SWI-Prolog and cplint code boxes -- see Figure~\ref{fig:codebox} -- by including their source directly in a Markdown file via a custom Prolog code listing or load the programme from an external code file. %
This plugin is based on SWISH\footnote{\url{https://swish.swi-prolog.org/}} -- SWI-Prolog for Sharing -- which is an online, interactive SWI-Prolog coding environment akin to Jupyter Notebooks~\cite{wielemaker2015swish}. %
The \texttt{sphinx-problog}\footnote{\url{https://github.com/simply-logical/sphinx-problog/}} extension, on the other hand, allows to embed interactive ProbLog code boxes, the implementation of which is based on the online execution environment underpinning the code examples published as part of the ProbLog website. %
Additionally, building these plugins prompted us to explore alternative technologies for composing (interactive) training resources, which in turn has inspired a prototype of a new publishing workflow in which multiple artefacts such as online documents, slides, and computational notebooks can be generated from a unified collection of source materials~\cite{sokol2021you}.%

\renewcommand\acksname{Acknowledgements}
\begin{acks}
The development of the \texttt{sphinx-prolog} and \texttt{sphinx-problog} Jupyter Book plugins was supported by the TAILOR Network -- an ICT-48 European AI Research Excellence Centre funded by EU Horizon 2020 research and innovation programme, grant agreement number 952215.%
\end{acks}

\bibliographystyle{ACM-Reference-Format}
\bibliography{paper}


\begin{thebibliography}{7}

%%% the Web address.

\ifx \showCODEN    \undefined \def \showCODEN     #1{\unskip}     \fi
\ifx \showDOI      \undefined \def \showDOI       #1{#1}\fi
\ifx \showISBNx    \undefined \def \showISBNx     #1{\unskip}     \fi
\ifx \showISBNxiii \undefined \def \showISBNxiii  #1{\unskip}     \fi
\ifx \showISSN     \undefined \def \showISSN      #1{\unskip}     \fi
\ifx \showLCCN     \undefined \def \showLCCN      #1{\unskip}     \fi
\ifx \shownote     \undefined \def \shownote      #1{#1}          \fi
\ifx \showarticletitle \undefined \def \showarticletitle #1{#1}   \fi
\ifx \showURL      \undefined \def \showURL       {\relax}        \fi
% invisible to TeX
\providecommand\bibfield[2]{#2}
\providecommand\bibinfo[2]{#2}
\providecommand\natexlab[1]{#1}
\providecommand\showeprint[2][]{arXiv:#2}

\bibitem[\protect\citeauthoryear{De~Raedt, Kimmig, and Toivonen}{De~Raedt
  et~al\mbox{.}}{2007}]%
        {deraedt2007problog}
\bibfield{author}{\bibinfo{person}{Luc De~Raedt}, \bibinfo{person}{Angelika
  Kimmig}, {and} \bibinfo{person}{Hannu Toivonen}.}
  \bibinfo{year}{2007}\natexlab{}.
\newblock \showarticletitle{{ProbLog}: {A} Probabilistic {Prolog} and Its
  Application in Link Discovery}. In \bibinfo{booktitle}{\emph{IJCAI}},
  Vol.~\bibinfo{volume}{7}. Hyderabad, \bibinfo{pages}{2462--2467}.
\newblock


\bibitem[\protect\citeauthoryear{{Executable Books Community}}{{Executable
  Books Community}}{2020}]%
        {jupyter2020book}
\bibfield{author}{\bibinfo{person}{{Executable Books Community}}.}
  \bibinfo{year}{2020}\natexlab{}.
\newblock \bibinfo{booktitle}{\emph{Jupyter Book}}.
\newblock
\urldef\tempurl%
\url{https://doi.org/10.5281/zenodo.4539666}
\showDOI{\tempurl}


\bibitem[\protect\citeauthoryear{Flach}{Flach}{1994}]%
        {flach1994simply}
\bibfield{author}{\bibinfo{person}{Peter Flach}.}
  \bibinfo{year}{1994}\natexlab{}.
\newblock \bibinfo{booktitle}{\emph{{S}imply {L}ogical -- {I}ntelligent
  {R}easoning by {E}xample}}.
\newblock \bibinfo{publisher}{John Wiley \& Sons, Inc.}
\newblock


\bibitem[\protect\citeauthoryear{Riguzzi}{Riguzzi}{2018}]%
        {riguzzi2018foundations}
\bibfield{author}{\bibinfo{person}{Fabrizio Riguzzi}.}
  \bibinfo{year}{2018}\natexlab{}.
\newblock \bibinfo{booktitle}{\emph{{F}oundations of {P}robabilistic {L}ogic
  {P}rogramming}}.
\newblock \bibinfo{publisher}{River Publishers}.
\newblock


\bibitem[\protect\citeauthoryear{Sokol and Flach}{Sokol and Flach}{2021}]%
        {sokol2021you}
\bibfield{author}{\bibinfo{person}{Kacper Sokol} {and} \bibinfo{person}{Peter
  Flach}.} \bibinfo{year}{2021}\natexlab{}.
\newblock \showarticletitle{You Only Write Thrice: {C}reating Documents,
  Computational Notebooks and Presentations From a Single Source}. In
  \bibinfo{booktitle}{\emph{Beyond static papers: {R}ethinking how we share
  scientific understanding in Machine Learning -- {ICLR} 2021 Workshop}}.
\newblock
\urldef\tempurl%
\url{https://doi.org/10.48550/arXiv.2107.06639}
\showDOI{\tempurl}


\bibitem[\protect\citeauthoryear{Wielemaker, Lager, and Riguzzi}{Wielemaker
  et~al\mbox{.}}{2015}]%
        {wielemaker2015swish}
\bibfield{author}{\bibinfo{person}{Jan Wielemaker},
  \bibinfo{person}{Torbj{\"o}rn Lager}, {and} \bibinfo{person}{Fabrizio
  Riguzzi}.} \bibinfo{year}{2015}\natexlab{}.
\newblock \showarticletitle{{SWISH}: {SWI-Prolog} for Sharing}. In
  \bibinfo{booktitle}{\emph{Proceedings of the International Workshop on
  User-Oriented Logic Programming (IULP 2015); 31st International Conference on
  Logic Programming (ICLP 2015)}}. \bibinfo{pages}{99--113}.
\newblock
\urldef\tempurl%
\url{https://doi.org/10.48550/arXiv.1511.00915}
\showDOI{\tempurl}


\bibitem[\protect\citeauthoryear{Wielemaker, Schrijvers, Triska, and
  Lager}{Wielemaker et~al\mbox{.}}{2012}]%
        {wielemaker2012swiprolog}
\bibfield{author}{\bibinfo{person}{Jan Wielemaker}, \bibinfo{person}{Tom
  Schrijvers}, \bibinfo{person}{Markus Triska}, {and}
  \bibinfo{person}{Torbj{\"o}rn Lager}.} \bibinfo{year}{2012}\natexlab{}.
\newblock \showarticletitle{{SWI-Prolog}}.
\newblock \bibinfo{journal}{\emph{Theory and Practice of Logic Programming}}
  \bibinfo{volume}{12}, \bibinfo{number}{1-2} (\bibinfo{year}{2012}),
  \bibinfo{pages}{67--96}.
\newblock
\urldef\tempurl%
\url{https://doi.org/10.1017/S1471068411000494}
\showDOI{\tempurl}


\end{thebibliography}

\end{document}